%% file: main.tex
\documentclass[sigconf]{acmart}
\input{macros}

\title{MobiDiff: Semantic-Aware Multi-Channel Discrete Diffusion for Human Mobility Data Generation}

\author{Rongchao Xu}
\affiliation{
  \institution{Florida State University}
  \city{Tallahassee, Florida}
  \country{USA}
}
\email{rx21a@fsu.edu}

\author{Lin Jiang}
\affiliation{
  \institution{Florida State University}
  \city{Tallahassee, Florida}
  \country{USA}
}
\email{lj23d@fsu.edu}

\author{Dahai Yu}
\affiliation{
  \institution{Florida State University}
  \city{Tallahassee, Florida}
  \country{USA}
}
\email{dahai.yu@fsu.edu}

\author{Ximiao Li}
\affiliation{
  \institution{Florida State University}
  \city{Tallahassee, Florida}
  \country{USA}
}
\email{xl24g@fsu.edu}

\author{Taichi Liu}
\affiliation{
  \institution{Rutgers University}
  \city{Piscataway, New Jersey}
  \country{USA}
}
\email{taichi.liu@rutgers.edu}

\author{Desheng Zhang}
\affiliation{
  \institution{Rutgers University}
  \city{Piscataway, New Jersey}
  \country{USA}
}
\email{desheng@cs.rutgers.edu}

\author{Yuan Tian}
\affiliation{
  \institution{University of California, Los Angeles}
  \city{Los Angeles, California
}
  \country{USA}
}
\email{yuant@ucla.edu}

\author{Guang Wang}
\authornote{Prof. Guang Wang is the corresponding author.}
\affiliation{
  \institution{Florida State University}
  \city{Tallahassee, Florida}
  \country{USA}
}
\email{guang@cs.fsu.edu}

\begin{document}

\begin{abstract}

\input{sections/01_abstract}
\end{abstract}



\maketitle

\section{Introduction}
\input{sections/02_introduction}

\section{Preliminaries}
\input{sections/04_preliminaries}

\section{Method}
\input{sections/05_method}

\section{Experiments}
\input{sections/06_experiments}

\section{Related Work}
\input{sections/03_related_work}

\section{Conclusion}
\input{sections/07_conclusion}

\bibliographystyle{ACM-Reference-Format}
\bibliography{main}

\end{document}

%% file: macros.tex
\newcommand{\methodname}{MobiDiff}

%% file: sections/01_abstract.tex
Human mobility data are essential for transportation optimization, urban planning, and resource allocation, yet real-world mobility data are costly to collect and difficult to share due to privacy concerns.
Recent diffusion-based methods have shown promise in synthesizing realistic mobility patterns, but they typically rely on continuous or latent spatio-temporal traces, limiting their ability to natively model discrete semantic events with explicit region, activity, time, and interval structures. To address this issue, we introduce \methodname{}, an end-to-end discrete diffusion framework that efficiently generates mobility data by directly denoising multi-channel semantic skeletons, avoiding the costly interpolation, latent trace construction, and coarse-to-fine realization pipelines widely used in existing diffusion-based methods. 
Specifically, \methodname{} decomposes each human check-in event into spatial, activity, and temporal channels, and employs structured event-, group-, and channel-level masking to jointly capture trajectory-level mobility patterns and within-event dependencies. We evaluate generation fidelity, privacy-preserving, and efficiency on three large-scale real-world datasets from Atlanta, Boston, and Seattle. 
Results show that \methodname{} effectively preserves trajectory length and temporal interval distributions while remaining competitive across broader mobility statistics; it is also much faster than state-of-the-art methods, e.g., 5.3$\times$ faster than GeoGen on average during inference. 
These findings suggest that discrete diffusion offers an interpretable and efficient framework for synthetic mobility data generation.

%% file: sections/02_introduction.tex
Human mobility data record where people go, when activities occur, and how semantic routines unfold across a day. 
Real mobility data are valuable for transportation optimization, urban planning, resource allocation, and ``what-if simulation''. However, they are difficult to collect at scale due to high collection costs and often difficult to share because they expose sensitive behavioral patterns \cite{kong2023mobility}. 
Synthetic data generation is therefore attractive as an efficient way to support large-scale data access while reducing direct exposure of raw human mobility traces. 
This need is reinforced by recent spatial-intelligence applications in ride-hailing, traffic forecasting, multivariate urban prediction, healthcare and energy demand prediction, and disaster-response analytics, where realistic spatio-temporal data are central to reliable modeling and decision support \cite{jiang2025hcride,li2024stsccl,yu2025uqgnn,yu2026health,yu2026trustenergy,yu2026energymamba,li2025typhoformer,jiang2025uncertainty,shen2025learning}. 

Recent advances in generative modeling have enabled synthetic mobility generation through diverse model families, including GAN-based, LLM-based, and diffusion-based methods.
GAN-based methods such as MoveSim \cite{feng2020movesim} generate mobility sequences through adversarial learning, but they often suffer from unstable training and mode collapse, limiting their ability to reliably capture diverse human mobility patterns.
LLM-based methods such as Geo-Llama \cite{li2024geollama} leverage large language models to model spatial-temporal mobility sequences and support mobility trajectory generation, but their autoregressive decoding and large model size can introduce substantial computational overhead for large-scale synthetic mobility data generation.
More recently, diffusion-based trajectory generators have improved sample quality by denoising continuous, latent, or staged spatio-temporal representations \cite{ho2020denoising}.
DiffTraj \cite{zhu2023difftraj} applies diffusion models to continuous GPS trajectory generation, AutoSTDiff \cite{xu2025autostdiff} introduces diffusion-based asynchronous trajectory generation, and GeoGen \cite{xu2026geogen} and SynHAT \cite{xu2026synhat} further develop coarse-to-fine diffusion pipelines for check-in and human mobility data synthesis.
However, existing diffusion-based mobility data generators still rely on continuous or latent trace construction, predefined temporal ranges or granularities, interpolation, or coarse-to-fine realization before producing the final mobility traces.
Consequently, their generative processes remain misaligned with the inherently discrete and structured nature of human mobility data, where each check-in event jointly encodes region, activity, time, and inter-event interval information.

Our key intuition is to directly model mobility trajectories as variable-length sequences of discrete semantic events rather than decomposing them into multiple continuous spatio-temporal traces with fixed lengths and intervals.
Discrete diffusion is well aligned with this representation because it operates directly on categorical mobility states, such as regions, activities, time bins, and interval bins, allowing the model to generate semantic activity skeletons in one stage without costly interpolation or coarse-to-fine realization. 
However, applying this idea to mobility data generation introduces two challenges.
First, each event contains tightly coupled spatial, semantic, and temporal factors, and modeling them independently can break realistic activity logic. 
Second, although these factors are represented as discrete tokens, they still carry numeric spatial-temporal meaning: region tokens reflect geographic proximity, absolute-time tokens encode daily periodicity, and gap-time tokens represent duration scales. 
A purely categorical treatment may therefore lose important mobility structure even when the generated skeletons appear valid.

To address these challenges, we propose \methodname{}, a semantic-aware multi-channel end-to-end discrete diffusion framework for large-scale human mobility data generation.
To capture the coupled factors within each activity event, \methodname{} represents every activity event as a multi-channel event composed of macro-region, micro-region, activity category, absolute-time, and gap-time tokens, and introduces structured masking at the event, group, and channel levels.
This masking strategy encourages the model to reconstruct missing spatial, semantic, and temporal components from both trajectory-level context and within-event dependencies.
To preserve the numeric meaning behind discrete tokens, \methodname{} further incorporates spatial- and temporal-aware token representations, where region tokens are associated with geographic coordinates and time/gap tokens encode periodicity and duration scales.
Consequently, \methodname{} can directly generate interpretable semantic activity skeletons while maintaining the spatial-temporal structure needed for realistic mobility trajectory synthesis.

We evaluate \methodname{} using Jensen--Shannon divergence over spatial, temporal, semantic, and aggregate trajectory statistics, together with efficiency comparisons against state-of-the-art diffusion-based baselines.
The results show that discrete diffusion is especially strong at preserving length and temporal interval behavior, remains competitive on broader mobility statistics, and offers a favorable efficiency profile, while spatial fidelity still leaves room for improvement on some city-metric combinations.

The main contributions of this work are summarized as follows:
\begin{itemize}
    \item We introduce a discrete diffusion perspective for human mobility data generation, formulating it as a one-stage end-to-end semantic activity skeleton denoising process rather than continuous trace synthesis.
    \item We propose \methodname{}, a semantic-aware multi-channel discrete diffusion framework that represents each activity event through spatial, semantic, and temporal tokens, and jointly models trajectory-level context and within-event dependencies through structured event-, group-, and channel-level masking.
    \item We conduct extensive experiments on three real-world human mobility datasets from Atlanta, Boston, and Seattle to evaluate generation fidelity, privacy preservation, and inference efficiency, demonstrating the effectiveness of \methodname{} and highlighting discrete diffusion as a promising future direction for human mobility data generation.
\end{itemize}

%% file: sections/04_preliminaries.tex
\subsection{Problem Statement}
\label{subsec:problem_statement}

\subsubsection{Human Mobility Trajectory}
A human mobility trajectory records a person's activity-level travel behavior as a sequence of sparse check-in events.
We denote a mobility trajectory as \(\mathcal{T}=[e_1,e_2,\ldots,e_L]\), where \(L\) is the trajectory length and each event \(e_i=(p_i,t_i)\) contains a point of interest (POI) \(p_i \in \mathcal{P}\) and a timestamp \(t_i\).
Here, \(\mathcal{P}\) is a finite POI set, and each POI is associated with metadata such as geographic coordinates and activity category.
Compared with continuous GPS traces, human mobility trajectories are sparse, irregularly sampled, and variable-length, since events are only observed when users visit meaningful activity locations.
The temporal interval between two consecutive events is defined as \(\Delta t_i=t_i-t_{i-1}\) for \(i>1\).

\subsubsection{Human Mobility Data Generation Problem}
Given a training set of real human mobility trajectories \(\mathcal{D}=\{\mathcal{T}^{(n)}\}_{n=1}^{N}\), the goal of human mobility data generation is to learn a generative model \(p_\theta\) with parameters \(\theta\) that can synthesize realistic trajectories \(\hat{\mathcal{T}}\sim p_\theta(\mathcal{T})\).
The generated trajectories are expected to preserve the spatial, semantic, temporal, and aggregate mobility patterns of real human mobility trajectories while avoiding direct reuse of individual real trajectories.

\subsection{Masked Discrete Diffusion}
\label{subsec:masked_discrete_diffusion}

Discrete diffusion models extend diffusion-based generative modeling to categorical data by defining a forward corruption process over discrete states and learning a reverse denoising process \cite{austin2021structured,hoogeboom2021argmax,campbell2022continuous}.
A common and effective variant is masked discrete diffusion, where corrupted tokens are replaced by a special absorbing state \texttt{[MASK]} and the model learns to recover the original clean tokens from partially masked inputs \cite{austin2021structured,shi2024simplified}.
Given a clean discrete sequence \(\mathbf{x}_0=[x_1,\ldots,x_n]\) of length \(n\), the forward corruption distribution \(q\) samples a corrupted sequence \(\mathbf{x}_t\) by independently masking tokens according to a timestep-dependent masking rate \(\alpha_t\):
\[
q(x_{t,j}\mid x_{0,j}) =
\begin{cases}
x_{0,j}, & \text{with probability } 1-\alpha_t,\\
\texttt{[MASK]}, & \text{with probability } \alpha_t.
\end{cases}
\]
As \(t\) increases, more tokens are masked, and the sequence becomes less informative.
The reverse model is trained to reconstruct the clean sequence from the corrupted sequence and the diffusion timestep:
\[
p_\theta(\mathbf{x}_0 \mid \mathbf{x}_t,t).
\]
The standard training objective minimizes the negative log-likelihood of clean tokens at masked positions:
\[
\mathcal{L}_{\mathrm{diff}}
=
-\mathbb{E}_{t,\mathbf{x}_0,\mathbf{x}_t}
\sum_{j:\,x_{t,j}=\texttt{[MASK]}}
\log p_\theta(x_{0,j}\mid \mathbf{x}_t,t).
\]

Masked discrete diffusion is well suited for human mobility data generation because mobility trajectories are naturally composed of categorical mobility states, such as POIs, activity categories, spatial regions, and temporal bins.
Unlike continuous diffusion, which adds Gaussian noise to coordinates or latent vectors, masked discrete diffusion operates directly in the symbolic space used by the generated trajectories.
This makes it possible to denoise corrupted activity-event tokens while preserving the discrete semantic structure of human mobility trajectories.

%% file: sections/05_method.tex
\begin{figure*}[t]
    \centering
    \includegraphics[width=\textwidth]{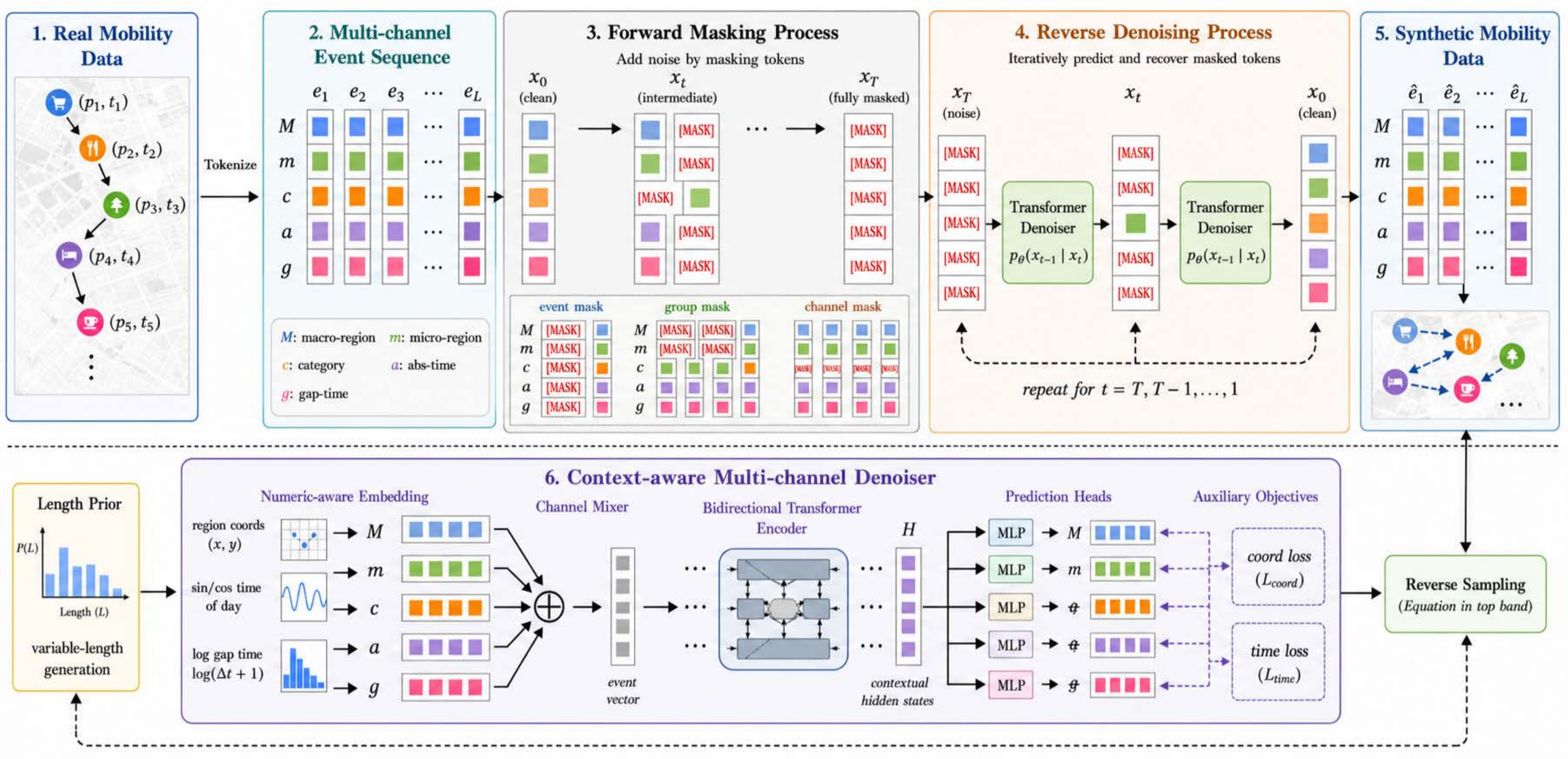}
    \caption{Overview of \methodname{}.
    The model first converts each human mobility trajectory into a variable-length sequence of multi-channel semantic events.
    It then learns a masked discrete diffusion process that reconstructs corrupted spatial, semantic, and temporal channels from trajectory context.}
    \label{fig:discretehat_framework}
\end{figure*}

\methodname{} is an end-to-end masked discrete diffusion framework for semantic human mobility data generation.
As shown in Figure~\ref{fig:discretehat_framework}, the framework first tokenizes each human mobility trajectory into a multi-channel semantic skeleton.
It then applies structured masked diffusion to corrupt event channels at different granularities.
A numeric-aware bidirectional denoiser reconstructs the masked channels from trajectory context.
Finally, the reverse sampler starts from a masked skeleton canvas and iteratively reveals high-confidence event tokens.
The following subsections describe the skeleton representation, forward masking process, denoiser, training objective, and reverse sampling procedure.

\subsection{Multi-channel Semantic Skeleton}
\label{subsec:semantic_skeleton}

Given a human mobility trajectory \(\mathcal{T}=[e_1,\ldots,e_L]\), we convert it into a skeleton sequence \(\mathbf{S}_0=[\mathbf{s}_1,\ldots,\mathbf{s}_L]\).
Each skeleton event is represented as
\[
\mathbf{s}_i=(z_i^M,z_i^m,z_i^c,z_i^a,z_i^g),
\]
where \(z_i^M\), \(z_i^m\), \(z_i^c\), \(z_i^a\), and \(z_i^g\) denote macro-region, micro-region, activity-category, absolute-time, and gap-time tokens, respectively.
The macro- and micro-region channels describe coarse and fine-grained spatial structure.
The category channel records the semantic activity type.
The absolute-time and gap-time channels encode daily rhythm and inter-event interval structure.
All channels share the same sequence length, and special tokens are reserved for padding, masking, and end-of-sequence markers.
This representation preserves the internal structure of check-in events while keeping the output discrete and inspectable.
Therefore, \methodname{} generates semantic skeleton trajectories rather than exact GPS coordinates, raw timestamps, or full POI-realized traces.

\subsection{Structured Masked Diffusion}
\label{subsec:structured_masking}

\methodname{} defines a masked forward process over the five-channel skeleton.
Let \(T\) denote the total number of diffusion steps.
For each training trajectory, we sample a diffusion step \(t\in\{1,\ldots,T\}\) and compute a masking ratio
\[
\rho_t = 1-\cos\left(\frac{\pi t}{2T}\right).
\]
A subset of valid event positions is selected according to \(\rho_t\).
Instead of masking each token independently, \methodname{} samples one of three masking granularities.
Event-level masking replaces all five channels of a selected event with \texttt{[MASK]}.
Group-level masking replaces a semantically meaningful channel group, such as spatial channels \(\{M,m\}\), temporal channels \(\{a,g\}\), or mixed groups \(\{M,m,c\}\) and \(\{c,a,g\}\).
Channel-level masking replaces one randomly selected channel of the selected events.

The corrupted skeleton at step \(t\) is denoted by \(\mathbf{S}_t\).
For a selected event \(i\) and channel \(r\in\{M,m,c,a,g\}\), the corrupted token \(z_{t,i}^{r}\) is defined by
\[
z_{t,i}^{r} =
\begin{cases}
\texttt{[MASK]}, & r \in \mathcal{A}_i,\\
z_{0,i}^{r}, & r \notin \mathcal{A}_i,
\end{cases}
\]
where \(z_{0,i}^{r}\) is the clean token and \(\mathcal{A}_i\) is the masked channel set determined by the sampled masking granularity.
This corruption process forces the denoiser to solve complementary reconstruction tasks.
Full-event masks require trajectory-level reasoning.
Group masks require cross-factor reasoning among spatial, semantic, and temporal variables.
Single-channel masks require within-event consistency recovery.

\subsection{Numeric-aware Denoiser}
\label{subsec:numeric_aware_denoiser}

The reverse model is a bidirectional Transformer denoiser \(p_\theta\), where \(\theta\) denotes all learnable model parameters.
For each channel \(r\in\{M,m,c,a,g\}\), \methodname{} first embeds the corrupted token \(z_{t,i}^{r}\) with a channel-specific embedding table.
For channels with numeric meaning, the token embedding is augmented with a small numeric projection:
\[
\mathbf{e}_i^{r}
=
\mathbf{E}_{r}(z_{t,i}^{r})
+
\phi_{r}(\mathbf{u}_i^{r}).
\]
Here, \(\mathbf{e}_i^{r}\) is the channel embedding, \(\mathbf{E}_r\) is the token-embedding table, \(\mathbf{u}_i^{r}\) is the numeric feature vector, and \(\phi_r\) is a channel-specific MLP projection.
For macro-region tokens, \(\mathbf{u}_i^{M}\) is the latitude-longitude coordinate of the region center.
For absolute-time tokens, \(\mathbf{u}_i^{a}\) contains the normalized bin center and its sinusoidal time-of-day features.
For gap-time tokens, \(\mathbf{u}_i^{g}\) is the log-transformed duration-bin center.
The micro-region and activity-category channels use pure token embeddings because their numeric side information is not assumed to be available.

The five channel embeddings of an event are concatenated and passed through an MLP channel mixer:
\[
\mathbf{v}_i = \psi([\mathbf{e}_i^M;\mathbf{e}_i^m;\mathbf{e}_i^c;\mathbf{e}_i^a;\mathbf{e}_i^g]).
\]
Here, \([\cdot;\cdot]\) denotes concatenation, \(\psi\) is the MLP channel mixer, and \(\mathbf{v}_i\) is the mixed event representation.
We then add sinusoidal positional encodings and process the event sequence with a Transformer encoder.
The encoder is bidirectional because the denoising task observes a partially corrupted trajectory rather than a causal prefix.
Let \(\mathbf{h}_i\) be the contextual hidden state at event position \(i\).
Each channel has its own prediction head:
\[
p_\theta(z_{0,i}^{r}\mid \mathbf{S}_t)
=
\mathrm{softmax}(\mathbf{W}_{r}\mathbf{h}_i+\mathbf{b}_{r}).
\]
Here, \(\mathbf{W}_{r}\) and \(\mathbf{b}_{r}\) are the output-head parameters for channel \(r\).
This factorized decoding preserves channel-specific vocabularies while sharing the same trajectory context.

\subsection{Training Objective}
\label{subsec:training_objective}

Training minimizes masked reconstruction loss over the corrupted channels.
Let \(\mathcal{M}_{r}\) be the set of valid positions where channel \(r\) is masked.
The diffusion loss is
\[
\mathcal{L}_{\mathrm{diff}}
=
\sum_{r\in\{M,m,c,a,g\}}
\lambda_r
\frac{1}{|\mathcal{M}_{r}|}
\sum_{i\in\mathcal{M}_{r}}
-
\log p_\theta(z_{0,i}^{r}\mid \mathbf{S}_t),
\]
where \(\lambda_r\) is an optional channel weight.
The implementation also uses inverse-frequency weighting for the activity-category channel to reduce the effect of long-tailed category frequencies.

To anchor discrete tokens to their numeric spatial-temporal meaning, \methodname{} adds auxiliary coordinate and time objectives.
The denoiser predicts a coordinate vector \(\hat{\mathbf{x}}_i\) and a normalized absolute-time value \(\hat{\tau}_i\) from \(\mathbf{h}_i\).
The corresponding targets are the macro-region coordinate \(\mathbf{x}_i\) and the normalized absolute-time bin center \(\tau_i\).
The auxiliary loss is
\[
\mathcal{L}_{\mathrm{aux}}
=
\lambda_{xy}
\sum_{i\in\mathcal{V}}
\|\hat{\mathbf{x}}_i-\mathbf{x}_i\|_2^2
+
\lambda_{\tau}
\sum_{i\in\mathcal{V}}
(\hat{\tau}_i-\tau_i)^2,
\]
where \(\mathcal{V}\) is the set of valid non-padding positions.
The coefficients \(\lambda_{xy}\) and \(\lambda_{\tau}\) control the weights of the coordinate and time objectives.
The full training objective is
\[
\mathcal{L}
=
\mathcal{L}_{\mathrm{diff}}
+
\mathcal{L}_{\mathrm{aux}}.
\]

\subsection{Reverse Sampling}
\label{subsec:reverse_sampling}

At inference time, \methodname{} first samples a trajectory length from the empirical training length distribution.
It then initializes all valid positions and channels with \texttt{[MASK]}.
Starting from step \(T\), the denoiser repeatedly predicts channel distributions for the current partially masked skeleton.
At each step, candidate tokens are sampled with the configured temperature and top-\(k\) rule.
The model computes a confidence score from the predicted probability of each sampled token.
For each channel, the highest-confidence masked positions are revealed according to the reverse cosine schedule.
This process continues until no valid positions remain masked.

The decoded output is a multi-channel semantic skeleton.
End-of-sequence and special tokens are removed during decoding.
When constructing valid generated human mobility trajectories, all channels are aligned to the same generated length.
For the semantic-consistency configuration used in the main comparison, the fine-grained location channel is treated as the anchor channel.
Train-derived dominant mappings from fine-grained location to macro-region and activity category are used to repair invalid or inconsistent macro/category tokens.
This step makes generated skeletons valid multi-channel check-in records without changing the core denoising model.

%% file: sections/06_experiments.tex
\subsection{Research Questions}
\label{subsec:research_questions}

Following the evaluation style of prior mobility-generation studies, we organize the experiments around the following research questions.
\begin{itemize}
    \item \textbf{RQ1:} How realistic are the generated human mobility trajectories in terms of spatial, temporal, semantic, and overall data distributions?
    \item \textbf{RQ2:} How efficient is the proposed discrete diffusion sampler during inference?
    \item \textbf{RQ3:} How useful are the generated human mobility data for downstream prediction tasks?
    \item \textbf{RQ4:} How much empirical training-trace exposure do the generated mobility trajectories exhibit?
\end{itemize}

\subsection{Experimental Setup}
\label{subsec:experimental_setup}

\subsubsection{Datasets}
We evaluate \methodname{} on three city-scale human mobility datasets collected from location-based check-in records.
Each dataset is converted into the same multi-channel semantic skeleton representation described in Section~\ref{subsec:semantic_skeleton}.
Table~\ref{tab:dataset_statistics} summarizes the dataset statistics.

\input{tables/tab_dataset_statistics}

For each city, models are trained on the training split and evaluated by comparing generated trajectories with the held-out test split.
Unless otherwise stated, all generated outputs are converted into the same trajectory format and evaluated with the same feature-extraction pipeline.
The primary comparison uses generated-versus-test metrics, while generated-versus-train metrics are used as a sanity check for overfitting and distribution drift.

\subsubsection{Baselines}
We compare \methodname{} with representative mobility-generation baselines.
\begin{itemize}
    \item GeoGen \cite{xu2026geogen} is a two-stage coarse-to-fine generator for fine-grained human mobility trajectories.
    \item SynHAT \cite{xu2026synhat} is a two-stage diffusion framework for synthesizing human mobility trajectories.
    \item MoveSim \cite{feng2020movesim} is a neural mobility simulator that learns human movement behavior from observed traces.
\end{itemize}

\subsubsection{Metrics}
We evaluate synthetic human mobility trajectories from three perspectives: fidelity, utility, and empirical exposure risk.
For fidelity, we follow the distribution-matching protocol used in synthetic mobility generation and compute Jensen--Shannon divergence (JSD) between generated and real feature distributions.
Lower JSD indicates better distributional similarity.
Given a real feature distribution \(P\) and a generated feature distribution \(Q\), we compute
\[
\mathrm{JSD}(P\|Q)
=
\frac{1}{2}\mathrm{KL}(P\|M)
+
\frac{1}{2}\mathrm{KL}(Q\|M),
\quad
M=\frac{1}{2}(P+Q),
\]
where \(\mathrm{KL}(\cdot\|\cdot)\) denotes Kullback--Leibler divergence.
The spatial group includes travel distance and movement radius.
The temporal group includes inter-event interval, trajectory length, and trajectory duration.
The semantic group includes POI diversity, POI entropy, category diversity, and category transition distributions.
The overall fidelity score is the mean JSD over all evaluated feature distributions.

For utility, we follow the standard downstream-evaluation protocol used by existing studies such as SynHAT \cite{xu2026synhat}.
Synthetic trajectories are used either as a replacement training set or as low-data augmentation together with a small real training subset.
We then evaluate next-event predictors on held-out real trajectories and report normalized Top-\(k\) ratios for macro-region, POI, and activity-category prediction.
Higher utility ratios indicate that the synthetic data better preserve task-relevant mobility patterns.

For empirical exposure risk, we measure the tight nearest-training overlap between each generated trajectory and its closest training trajectory.
The tight screen counts event-level matches under a spatial threshold of 0.2 km and a temporal threshold of 30 minutes.
Lower overlap indicates less direct reuse of training traces.
This metric is an empirical memorization diagnostic rather than a formal privacy guarantee.

\subsection{RQ1: Generation Fidelity}
\label{subsec:rq1_fidelity}

Table~\ref{tab:fidelity_baselines} reports generated-versus-test JSD on selected component metrics related to representative spatial, temporal, and semantic distributions.
The selected metrics include distance and radius for spatial structure, interval, length, and duration for temporal structure, and POI diversity and POI entropy for fine-grained semantic coverage.
The Avg. column is the mean of the seven selected metrics, so lower values indicate better fidelity.

\input{tables/tab_fidelity_baselines}

\methodname{} obtains the lowest selected average JSD in Seattle and the second-lowest selected average JSD in Atlanta and Boston.
Averaged over the three cities, its selected average JSD is 0.084, compared with 0.216 for GeoGen \cite{xu2026geogen}, 0.174 for SynHAT \cite{xu2026synhat}, and 0.161 for MoveSim \cite{feng2020movesim}.
This corresponds to a 61\% reduction relative to GeoGen, a 52\% reduction relative to SynHAT, and a 48\% reduction relative to MoveSim on these selected metrics.
The improvement is most pronounced for temporal behavior.
Across interval, length, and duration, \methodname{} has an average JSD of 0.039, while GeoGen, SynHAT, and MoveSim have averages of 0.290, 0.263, and 0.255, respectively.
This supports the central design choice of representing absolute time and gap time as explicit channels instead of treating time as an external post-processing variable.

The semantic results are also consistent with the multi-channel skeleton design.
Across POI diversity and POI entropy, \methodname{} has an average JSD of 0.107.
This is lower than GeoGen at 0.251, SynHAT at 0.181, and MoveSim at 0.142.
At the same time, the spatial columns show that \methodname{} is not uniformly best.
SynHAT and MoveSim better preserve distance or radius in several city-metric combinations, and the Boston radius is a clear weakness for \methodname{}.
The fidelity conclusion is therefore measured: discrete masked denoising gives strong temporal and selected semantic fidelity, but spatial spread still needs refinement.

\subsection{RQ2: Inference Efficiency}
\label{subsec:rq2_efficiency}

Table~\ref{tab:efficiency} reports inference throughput and peak GPU memory.
The benchmark uses a 256-trajectory inference probe with three repeated measurements after warmup.
For multi-stage diffusion baselines, the reported throughput measures the dominant diffusion sampling stage used in the adapted pipeline.

\input{tables/tab_efficiency}

\methodname{} generates 19.82 trajectories per second on average.
This is 5.3\(\times\) faster than GeoGen \cite{xu2026geogen} and 1.9\(\times\) faster than SynHAT \cite{xu2026synhat} in the measured inference probe.
The speedup follows from denoising semantic skeletons in one discrete sampling pipeline rather than using a coarse-to-fine diffusion realization process.
However, \methodname{} is not the fastest generator overall.
MoveSim reaches 33.41 trajectories per second.
\methodname{} also uses 1168 MB of peak GPU memory, which is higher than all baselines in Table~\ref{tab:efficiency}.
Thus, the efficiency claim is diffusion-specific: \methodname{} substantially reduces sampling time relative to two-stage diffusion baselines, but lightweight non-diffusion generators remain faster and more memory-efficient.

\subsection{RQ3: Downstream Utility}
\label{subsec:rq3_utility}

We evaluate downstream utility by training next-event prediction models on synthetic data and testing them on real held-out trajectories.
Tables~\ref{tab:utility_replacement} and~\ref{tab:utility_augmentation} report normalized Top-\(k\) utility ratios under synthetic replacement and low-data augmentation.
All ratios are normalized by the corresponding real-data training performance, so higher values indicate stronger downstream usefulness.

\input{tables/tab_utility_replacement}

In the synthetic-replacement setting, \methodname{} reaches city-averaged Macro Top-5, POI Top-20, and Category Top-5 ratios of 0.654, 0.480, and 0.880, respectively.
Its POI replacement utility is about 5.0\(\times\) that of GeoGen and 1.9\(\times\) that of SynHAT.
This indicates that directly denoising semantic skeleton events preserves more fine-grained POI signal than the two diffusion baselines in the replacement setting.
However, \methodname{} trails MoveSim on POI replacement, where it reaches an average of 0.688.
The macro-region and category columns show the same limitation: \methodname{} is useful, but autoregressive or simulation-style training remains better aligned with next-event prediction.

\input{tables/tab_utility_augmentation}

In the low-data augmentation setting, \methodname{} reaches city-averaged Macro Top-5, POI Top-20, and Category Top-5 ratios of 0.907, 0.665, and 0.962.
The POI augmentation ratio improves over GeoGen by 10\% and over SynHAT by 28\%.
The category ratio is also close to SynHAT and stronger than GeoGen, although MoveSim remains the strongest utility baseline.
These results suggest that \methodname{} produces semantic skeletons that can supplement limited real data, especially for POI and category learning.
They also expose a limitation of masked denoising: the training objective reconstructs corrupted events from bidirectional context, while downstream utility is evaluated through next-event prediction.
Future variants may need utility-aware objectives that improve recommendation value without making samples more train-like.

\subsection{RQ4: Empirical Exposure Risk}
\label{subsec:rq4_exposure}

We finally evaluate empirical exposure risk using the tight nearest-training overlap screen.
This diagnostic compares each generated trajectory with its nearest training trajectory under a 0.2 km spatial tolerance and a 30-minute temporal tolerance.
Lower values indicate lower empirical memorization risk, but the metric is not a formal differential-privacy guarantee.
Table~\ref{tab:privacy} reports upper-tail overlap, mean overlap, and calibration ratio for each city.

\input{tables/tab_privacy}

\methodname{} has a city-averaged \(P_{95}\) overlap of 0.783 and a city-averaged mean overlap of 0.660.
They are also lower than MoveSim, whose \(P_{95}\) is 1.000 and whose mean overlap is 0.786.
Relative to these two high-utility baselines, \methodname{} reduces mean overlap by 25\% and 16\%, respectively.
However, GeoGen and SynHAT have lower empirical overlap in several city-level diagnostics.
GeoGen has an average mean overlap of 0.437, and SynHAT has an average mean overlap of 0.463.
The correct interpretation is therefore not that \methodname{} solves privacy, but that it avoids the saturated overlap behavior of the strongest non-diffusion utility baselines.

%% file: tables/tab_dataset_statistics.tex
\begin{table}[t]
\centering
\caption{Dataset statistics.}
\label{tab:dataset_statistics}
\small
\begin{tabular}{lrrrrr}
\hline
City & \#Users & \#POIs & \#Check-ins & \#Trajectories & Avg. length \\
\hline
Atlanta & 174,787 & 5,178 & 2,324,746 & 289,175 & 8.04 \\
Boston  & 72,296  & 2,843 & 1,075,084 & 126,366 & 8.51 \\
Seattle & 113,509 & 4,850 & 1,871,657 & 215,128 & 8.70 \\
\hline
\end{tabular}
\end{table}

%% file: tables/tab_fidelity_baselines.tex
\begin{table*}[t]
\centering
\caption{Fidelity evaluation on selected generated-versus-test JSD metrics. Best values are in bold, second-best values are underlined, and the \methodname{} row is highlighted for emphasis.}
\label{tab:fidelity_baselines}
\scriptsize
\setlength{\tabcolsep}{3pt}
\renewcommand{\arraystretch}{1.08}
\resizebox{\textwidth}{!}{%
\begin{tabular}{l|rrrr|rrrr|rrrr}
\hline
\textbf{Method}
& \multicolumn{4}{c|}{\textbf{Atlanta}}
& \multicolumn{4}{c|}{\textbf{Boston}}
& \multicolumn{4}{c}{\textbf{Seattle}} \\
\cline{2-13}
& Distance & Radius & Interval & Length
& Distance & Radius & Interval & Length
& Distance & Radius & Interval & Length \\
\hline
GeoGen
& 0.1082 & 0.1070 & 0.0779 & 0.4009
& \underline{0.0015} & 0.0015 & 0.0709 & 0.3566
& 0.0659 & \underline{0.0921} & \underline{0.0878} & 0.3565 \\
SynHAT
& \textbf{0.0017} & \underline{0.0307} & \underline{0.0769} & \underline{0.2426}
& \textbf{0.0006} & \underline{0.0007} & \underline{0.0584} & \underline{0.1153}
& 0.0878 & \textbf{0.0829} & 0.0996 & 0.6408 \\
MoveSim
& \underline{0.0053} & \textbf{0.0020} & 0.1409 & 0.2934
& \underline{0.0015} & \textbf{0.0006} & 0.1238 & 0.1993
& \textbf{0.0385} & 0.1090 & 0.1342 & \underline{0.1929} \\
\textbf{\methodname{}}
& 0.0425 & 0.1086 & \textbf{0.0355} & \textbf{0.0353}
& 0.1112 & 0.1996 & \textbf{0.0186} & \textbf{0.0138}
& \underline{0.0529} & 0.1274 & \textbf{0.0279} & \textbf{0.0342} \\
\hline
\multicolumn{13}{c}{} \\
\hline
& \multicolumn{4}{c|}{\textbf{Atlanta}}
& \multicolumn{4}{c|}{\textbf{Boston}}
& \multicolumn{4}{c}{\textbf{Seattle}} \\
\cline{2-13}
& Duration & POI div. & POI ent. & Avg.
& Duration & POI div. & POI ent. & Avg.
& Duration & POI div. & POI ent. & Avg. \\
\hline
GeoGen
& 0.4049 & 0.3948 & 0.2779 & 0.253
& 0.4153 & 0.2305 & \underline{0.1173} & 0.171
& 0.4475 & 0.3019 & 0.2196 & 0.224 \\
SynHAT
& \underline{0.3412} & \underline{0.1387} & 0.2029 & \underline{0.148}
& \underline{0.3287} & \underline{0.0669} & \textbf{0.0789} & \underline{0.093}
& 0.4595 & 0.3192 & 0.2826 & 0.282 \\
MoveSim
& 0.4325 & 0.1862 & \underline{0.1729} & 0.176
& 0.4055 & 0.1364 & 0.1329 & 0.143
& \underline{0.4128} & \underline{0.1541} & \textbf{0.1081} & \underline{0.164} \\
\textbf{\methodname{}}
& \textbf{0.1235} & \textbf{0.0695} & \textbf{0.0830} & \textbf{0.071}
& \textbf{0.0350} & \textbf{0.0633} & 0.1427 & \textbf{0.083}
& \textbf{0.0258} & \textbf{0.0963} & \underline{0.1862} & \textbf{0.079} \\
\hline
\end{tabular}
}
\end{table*}

%% file: tables/tab_efficiency.tex
\begin{table}[t]
\centering
\caption{Inference efficiency. Throughput is measured in generated trajectories per second, and memory is peak GPU memory in megabytes.}
\label{tab:efficiency}
\footnotesize
\resizebox{\columnwidth}{!}{%
\begin{tabular}{lrrrrr}
\hline
Method & ATL & BOS & SEA & Mean & Mem. \\
\hline
GeoGen & 3.27 & 4.33 & 3.64 & 3.75 & 346 \\
SynHAT & 10.22 & 10.26 & 10.18 & 10.22 & 332 \\
MoveSim & 26.47 & 42.78 & 30.98 & 33.41 & 744 \\
\methodname{} & 18.27 & 21.62 & 19.58 & 19.82 & 1168 \\
\hline
\end{tabular}
}
\end{table}

%% file: tables/tab_utility_replacement.tex
\begin{table*}[t]
\centering
\caption{City-level downstream utility evaluation under synthetic replacement. Values are normalized utility ratios, where higher is better. Best values are in bold and second-best values are underlined.}
\label{tab:utility_replacement}
\scriptsize
\setlength{\tabcolsep}{4.2pt}
\renewcommand{\arraystretch}{1.08}
\resizebox{\textwidth}{!}{%
\begin{tabular}{lccc ccc ccc}
\toprule
\textbf{Method}
& \multicolumn{3}{c}{\textbf{Atlanta}}
& \multicolumn{3}{c}{\textbf{Boston}}
& \multicolumn{3}{c}{\textbf{Seattle}} \\
\cmidrule(lr){2-4}
\cmidrule(lr){5-7}
\cmidrule(lr){8-10}
& Macro Top-5 & POI Top-20 & Cat. Top-5
& Macro Top-5 & POI Top-20 & Cat. Top-5
& Macro Top-5 & POI Top-20 & Cat. Top-5 \\
\midrule
GeoGen
& 0.347 & 0.142 & 0.634
& 0.814 & 0.077 & 0.831
& 0.743 & 0.068 & 0.609 \\
SynHAT
& \textbf{0.920} & 0.275 & \underline{0.886}
& \textbf{0.969} & 0.311 & \underline{0.977}
& \underline{0.944} & 0.185 & \underline{0.876} \\
MoveSim
& \underline{0.875} & \textbf{0.666} & \textbf{0.954}
& \underline{0.960} & \textbf{0.783} & \textbf{0.986}
& \textbf{0.952} & \textbf{0.615} & \textbf{0.933} \\
\textbf{\methodname{}}
& 0.597 & \underline{0.435} & 0.868
& 0.734 & \underline{0.541} & 0.939
& 0.630 & \underline{0.465} & 0.834 \\
\bottomrule
\end{tabular}
}
\end{table*}

%% file: tables/tab_utility_augmentation.tex
\begin{table*}[t]
\centering
\caption{City-level downstream utility evaluation under low-data augmentation. Values are normalized utility ratios, where higher is better. Best values are in bold and second-best values are underlined.}
\label{tab:utility_augmentation}
\scriptsize
\setlength{\tabcolsep}{4.2pt}
\renewcommand{\arraystretch}{1.08}
\resizebox{\textwidth}{!}{%
\begin{tabular}{lccc ccc ccc}
\toprule
\textbf{Method}
& \multicolumn{3}{c}{\textbf{Atlanta}}
& \multicolumn{3}{c}{\textbf{Boston}}
& \multicolumn{3}{c}{\textbf{Seattle}} \\
\cmidrule(lr){2-4}
\cmidrule(lr){5-7}
\cmidrule(lr){8-10}
& Macro Top-5 & POI Top-20 & Cat. Top-5
& Macro Top-5 & POI Top-20 & Cat. Top-5
& Macro Top-5 & POI Top-20 & Cat. Top-5 \\
\midrule
GeoGen
& 0.841 & 0.559 & 0.930
& 0.960 & \underline{0.701} & 0.974
& 0.938 & \underline{0.590} & 0.945 \\
SynHAT
& \textbf{0.949} & 0.471 & 0.928
& \textbf{0.982} & 0.513 & \underline{0.981}
& \underline{0.967} & 0.558 & \underline{0.967} \\
MoveSim
& \underline{0.934} & \textbf{0.742} & \textbf{0.961}
& \underline{0.976} & \textbf{0.892} & \textbf{0.993}
& \textbf{0.981} & \textbf{0.852} & \textbf{0.980} \\
\textbf{\methodname{}}
& 0.901 & \underline{0.741} & \underline{0.943}
& 0.921 & 0.669 & 0.979
& 0.899 & 0.584 & 0.964 \\
\bottomrule
\end{tabular}
}
\end{table*}

%% file: tables/tab_privacy.tex
\begin{table*}[t]
\centering
\caption{City-level empirical exposure-risk evaluation under the tight nearest-training-overlap screen. Lower values indicate lower exposure risk. Best values are in bold and second-best values are underlined.}
\label{tab:privacy}
\scriptsize
\setlength{\tabcolsep}{8pt}
\renewcommand{\arraystretch}{1.05}
\resizebox{\textwidth}{!}{%
\begin{tabular}{lccc ccc ccc}
\toprule
\textbf{Method}
& \multicolumn{3}{c}{\textbf{Atlanta}}
& \multicolumn{3}{c}{\textbf{Boston}}
& \multicolumn{3}{c}{\textbf{Seattle}} \\
\cmidrule(lr){2-4}
\cmidrule(lr){5-7}
\cmidrule(lr){8-10}
& $P_{95}$ & Mean & Calib.
& $P_{95}$ & Mean & Calib.
& $P_{95}$ & Mean & Calib. \\
\midrule
GeoGen
& \underline{0.793} & \underline{0.485} & \underline{0.622}
& \textbf{0.556} & \textbf{0.394} & \textbf{0.446}
& \textbf{0.619} & \underline{0.432} & \underline{0.500} \\
SynHAT
& \textbf{0.690} & \textbf{0.423} & \textbf{0.544}
& 1.000 & \underline{0.577} & \underline{0.652}
& \underline{0.625} & \textbf{0.390} & \textbf{0.452} \\
MoveSim
& 1.000 & 0.735 & 0.944
& 1.000 & 0.822 & 0.929
& 1.000 & 0.801 & 0.928 \\
\textbf{\methodname{}}
& 0.800 & 0.667 & 0.857
& \underline{0.800} & 0.649 & 0.732
& 0.750 & 0.663 & 0.765 \\
\bottomrule
\end{tabular}
}
\end{table*}

%% file: sections/03_related_work.tex
\subsection{Synthetic Mobility Data Generation}

Synthetic mobility generation has been studied as a way to support mobility simulation, data sharing, and downstream model development when real traces are costly or sensitive to release \cite{kong2023mobility}.
This motivation is consistent with recent spatial-intelligence studies that rely on spatio-temporal traces or flows for traffic forecasting, ride-hailing, visit prediction, energy usage prediction, and disaster-response decisions \cite{li2024stsccl,jiang2025hcride,yu2025uqgnn,yu2026health,yu2026trustenergy,yu2026energymamba,jiang2025uncertainty,shen2025learning}.
Early neural approaches often formulate trajectory generation as sequential token generation.
SeqGAN \cite{yu2017seqgan} introduces adversarial sequence generation with policy gradients, and it can be adapted to discrete mobility tokens.
MoveSim \cite{feng2020movesim} learns to simulate human mobility behavior and provides a strong non-diffusion baseline for downstream utility.
Recent large-model approaches such as Geo-Llama \cite{li2024geollama} further explore language-model priors for mobility generation under spatio-temporal constraints.
These methods are effective for sequential generation, but they do not explicitly denoise the coupled spatial, semantic, and temporal channels inside each human mobility activity event.
In contrast, \methodname{} treats each check-in as a multi-channel semantic event and learns to reconstruct corrupted event components from full trajectory context.

\subsection{Diffusion Models for Trajectory Generation}

Diffusion models have become a powerful class of generative models after the success of denoising diffusion probabilistic models \cite{ho2020denoising}.
Trajectory-generation methods have adapted this idea to mobility data through continuous, latent, or staged spatio-temporal representations.
DiffTraj \cite{zhu2023difftraj} applies diffusion probabilistic modeling to continuous GPS trajectories.
AutoSTDiff \cite{xu2025autostdiff} introduces autoregressive spatio-temporal denoising for asynchronous trajectory generation.
GeoGen \cite{xu2026geogen} uses a two-stage coarse-to-fine framework for fine-grained location-based social network trajectory generation.
SynHAT \cite{xu2026synhat} further develops a two-stage coarse-to-fine diffusion framework for synthesizing human activity traces.
These studies show that diffusion can improve synthetic mobility quality, but they often rely on continuous locations, predefined temporal grids, interpolation, or coarse-to-fine realization.
\methodname{} differs by applying discrete diffusion directly to semantic mobility data skeletons, which avoids a separate continuous or latent trace construction step.

\subsection{Discrete Diffusion and Masked Denoising}

Discrete diffusion models extend diffusion-style generation from continuous vectors to categorical state spaces.
Structured denoising diffusion models define corruption and reverse processes over discrete variables \cite{austin2021structured}.
Argmax flows and multinomial diffusion provide another route for learning categorical distributions with diffusion-inspired objectives \cite{hoogeboom2021argmax}.
Continuous-time discrete denoising frameworks further generalize categorical diffusion processes \cite{campbell2022continuous}.
More recent masked-diffusion formulations simplify discrete generation by reconstructing masked tokens rather than adding Gaussian noise \cite{shi2024simplified}.
These methods motivate our masked denoising formulation, but human mobility data generation introduces additional structure that generic discrete diffusion does not address.
Each event couples spatial region, fine-grained location, activity category, absolute time, and gap time.
Moreover, several categorical tokens carry numeric meaning, such as geographic proximity, temporal periodicity, and duration scale.
\methodname{} therefore combines masked discrete diffusion with multi-channel event tokenization and numeric-aware embeddings, making discrete diffusion better aligned with semantic human mobility data generation.

%% file: sections/07_conclusion.tex
We presented \methodname{}, a semantic-aware multi-channel discrete diffusion framework for human mobility data generation.
Instead of generating continuous trajectories or latent mobility states, \methodname{} models human mobility trajectories as variable-length sequences of semantic skeleton events.
Each event jointly represents macro-region, micro-region, activity category, absolute-time bin, and gap-time bin.
Structured event-, group-, and channel-level masking encourages the denoiser to recover spatial, semantic, and temporal dependencies from trajectory context.
Numeric-aware embeddings further preserve geographic and temporal structure behind discrete tokens.
Experiments on three city-scale human mobility datasets from Atlanta, Boston, and Seattle show that \methodname{} is especially effective at preserving temporal and selected semantic distributions.
It also samples substantially faster than state-of-the-art two-stage diffusion baselines in the measured inference probe.
The downstream utility and exposure-risk results are more nuanced.
\methodname{} provides useful POI and category signals, especially under low-data augmentation, but autoregressive and simulation baselines remain stronger for some next-event prediction tasks.
Its empirical exposure risk is lower than SeqGAN and MoveSim under the tight nearest-training-overlap screen, but GeoGen and SynHAT achieve lower overlap on several diagnostics.
These results suggest that discrete diffusion is a promising and interpretable direction for semantic-aware human mobility data generation, while spatial fidelity, utility-aware denoising, and stronger exposure control remain important future work.

\section{ACKNOWLEDGMENT}
This work is partially supported by the National Science Foundation under Grants 2411151, 2411152, 2411153, the National Artificial Intelligence Research Resource (NAIRR) 240332, and
FSU/AWS Computer Support Seed Fund.